\documentclass[letterpaper, 10 pt, conference]{ieeeconf}  

\IEEEoverridecommandlockouts                              

\usepackage{amsmath} 
\usepackage{multirow}
\usepackage{makecell}
\usepackage{array}
\usepackage{graphicx}
\usepackage{stfloats}
\usepackage[
    singlelinecheck=false
]{caption}
\title{\LARGE \bf
Exploiting latent representation of sparse semantic layers for improved short-term motion prediction with Capsule Networks
}

\author{Albert Dulian and John C. Murray
\thanks{The authors are with the Department of Computer Science and Technology, The University of Hull, Kingston Upon Hull, United Kingdom 
{\tt\small A.Dulian-2013@hull.ac.uk, John.Murray@hull.ac.uk}}}

\begin{document}
\maketitle
\thispagestyle{empty}
\pagestyle{empty}
\begin{abstract}
As urban environments manifest high levels of complexity it is of vital importance that safety systems embedded within autonomous vehicles (AVs) are able to accurately anticipate short-term future motion of nearby agents. This problem can be further understood as generating a sequence of coordinates describing the future motion of the tracked agent. Various proposed approaches demonstrate significant benefits of using a rasterised top-down image of the road, with a combination of Convolutional Neural Networks (CNNs), for extraction of relevant features that define the road structure (eg. driveable areas, lanes, walkways). In contrast, this paper explores use of Capsule Networks (CapsNets) in the context of learning a hierarchical representation of sparse semantic layers corresponding to small regions of the High-Definition (HD) map. Each region of the map is dismantled into separate geometrical layers that are extracted with respect to the agent's current position. By using an architecture based on CapsNets the model is able to retain hierarchical relationships between detected features within images whilst also preventing loss of spatial data often caused by the pooling operation. We train and evaluate our model on publicly available dataset nuTonomy scenes and compare it to recently published methods. We show that our model achieves significant improvement over recently published works on deterministic prediction, whilst drastically reducing the overall size of the network.

\end{abstract}
\section{INTRODUCTION}
Autonomous vehicles (AVs) are expected to provide a safe and robust transportation solution in diverse and often highly uncertain surroundings. One of the key aspects required to achieve this technology is providing AVs with an ability to forecast short-term (e.g. 4 seconds) movements of other vehicles within close proximity. This can allow for further reasoning with regards to risk-free path planning and manoeuvre execution. Nonetheless, short-term trajectory prediction is not a trivial task due to the inherently difficult derivation of representative context cues with regards to agents and the environment.

Recent techniques for motion prediction have focused on improving the accuracy of predicted trajectories by incorporating rendered images of High-Definition (HD) maps in a top-down manner where various road attributes are specifically coded with different colours \cite{choi2019drogon, cui2019multimodal, marchetti2020mantra, phan2020covernet}. The rasterised map is then used as an input to a \textit{Convolutional Neural Network} (CNN) \cite{lecun1990handwritten} to encode global spatial context of the scene and combined with an agent's observed motion data for further inference.
\begin{figure}[t]
  \centering
   \includegraphics[scale=.55]{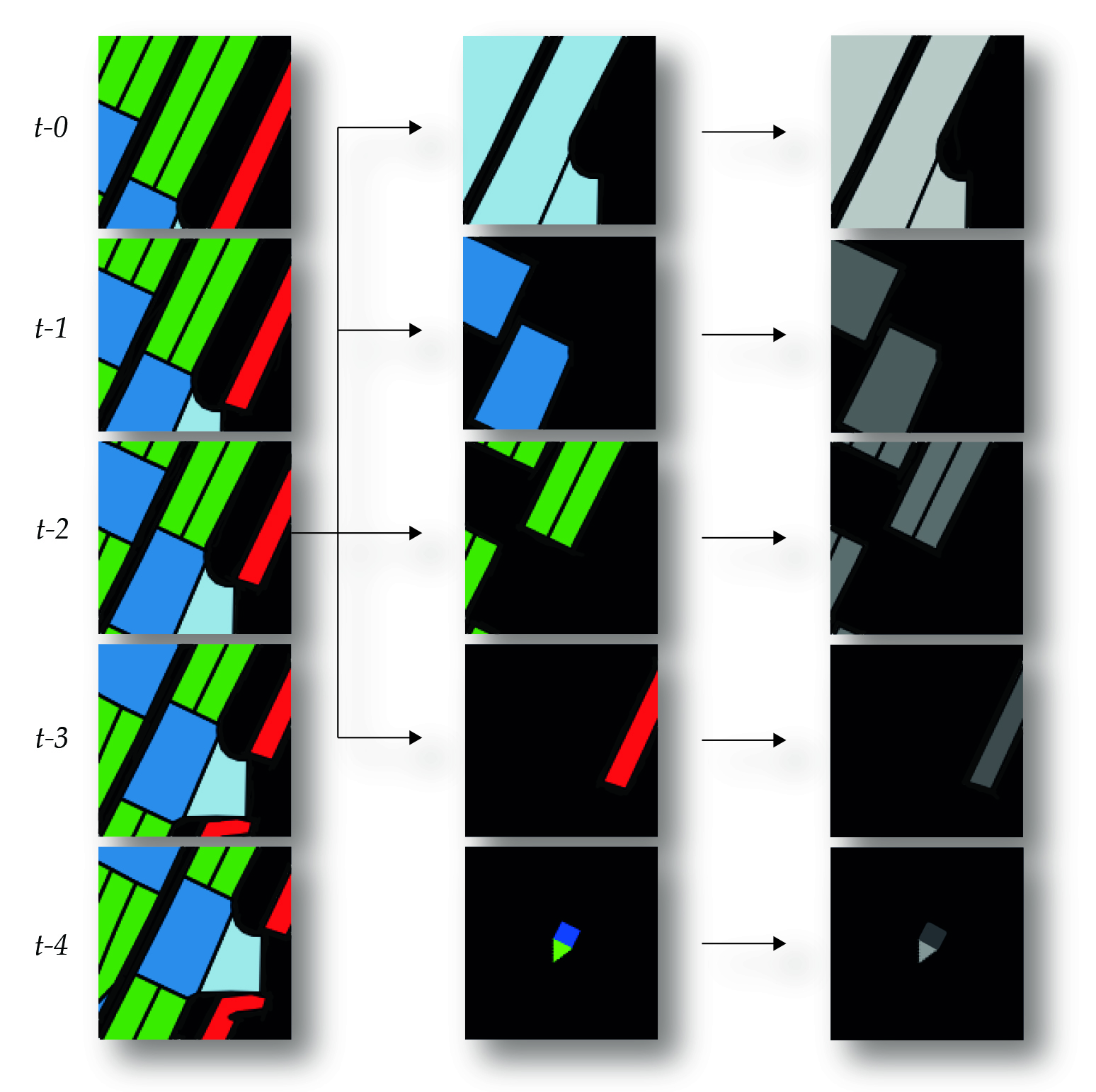}
  \caption{Disentanglement process of a map chunk at time $t-2$. The map is turned into several geometrical layers each representing a single part of the whole map. In addition, we draw an agent (bottom middle and right images) with its origin corresponding to the position on the chunk of interest.}
  \label{fig:map_dis}
\end{figure}

Yet, despite good results and definite advantages of incorporating map data into the network, there are still numerous potential limitations that might arise from these methods, effectively restricting the accuracy of the predicted motion. First of all, the scene context is extracted globally i.e. for each tracked agent an identical, single rasterized image is used to encode spatial information. We argue that this approach leads to a great loss of valuable data that can otherwise be associated with an individual agent, by extracting small local regions of the map. These regions correspond to the agent's current position (Fig. \ref{fig:map_dis}), thus enabling the model to learn a more representative latent space with respect to each tracked entity. We also believe that incorporating local map regions characterised by several semantic layers could potentially lead to the trajectory being predicted with respect to physical boundaries of the road.

In addition, a typical approach in which previously mentioned methods encode the image map data is by employing a variation of a CNN architecture e.g. \textit{ResNet-50} \cite{he2016deep} to extract more complex features with each consecutive convolutional layer. Although over the past number of years CNNs have achieved state-of-the-art performance in several computer vision tasks such as object classification \cite{touvron2020fixing}, detection \cite{acharya2020rodeo}, and semantic segmentation \cite{tao2020hierarchical}, these type of networks still exhibit two important drawbacks \cite{goodfellow2016deep}:
\begin{enumerate}
    \item \textit{Equivariance} - CNNs are not naturally equivariant to other transformations but translation (parameter sharing). For instance, a slight rotation of an object within the image will often cause relevant neurons to not fire and thus fail when detecting salient features.
    \item \textit{Local Invariance} - CNNs achieve local transformation invariance e.g. translation and rotation through pooling operation by taking maximum value (max-pooling) from a small window of former convolution layer's output. Although pooling operation might not have significant drawbacks if the aim is to solely detect whether the object is present or not, it does ignore hierarchical relationship between features and leads to loss of large volumes of spatial data.
\end{enumerate}
To tackle these issues Hinton \textit{et al.} \cite{hinton2011transforming} proposed a novel type of neural network known as \textit{Capsule Network} (CapsNet) that implements an idea of using capsules (locally invariant group of neurons) to learn various properties (e.g. pose) of the same object and encode them in an output vector whose length corresponds to the probability of that object being present. Encoded parameters can be further conceptualised as an object's instantiation parameters that enable the model to learn more robust and equivariant representation of features with respect to change in viewpoint. Furthermore, Sabour \textit{et al.} \cite{sabour2017dynamic} introduced the \textit{routing-by-agreement} mechanism by which capsules from lower levels decide which of their output vectors should be sent to higher level capsules. In essence, output vectors from lower level capsules are used to predict output of higher level capsules, predictions are then compared with actual outputs to iteratively compute "agreement" (cosine similarity) between lower and higher capsules. For instance, the mere presence of a nose or eyes (lower level capsules) should not be a sole indicator that the face (higher level capsules) exists within an image, a hierarchical relationship (e.g. rotation) between low and high level features should have a high impact on the final prediction. 

We are strongly motivated by the advantages offered by CapsNets and believe that this type of network is ideal for building a well representative latent space of map data, whose hierarchical relationship between features is of great importance. We therefore propose an architecture where a standard CNN backbone is replaced with Capsule layers. To the best of our knowledge this is the first piece of work that proposes use of CapsNets in the context of short-term motion prediction for autonomous vehicles. Through empirical evaluation we demonstrate the benefits of using a CapsNet based network. To summarise, our contributions are:
\begin{itemize}
    \item We demonstrate the first use of Capsule Networks with respect to the task of predicting short-term future motion of vehicles in complex environments.
    \item We propose the use of a local hierarchical semantic layers as opposed to a global rasterized top-down view of the environment. This aids with creating a more meaningful latent representation of the tracked agent.
    \item We show that a network with CapsNet feature extractor can outperform a CNN encoder based on e.g. ResNet architecture, whilst also significantly reducing number of parameters within the network.
    \item We examine and compare the performance of our proposed approach against state-of-the-art methods on a publicly available dataset
\end{itemize}

\section{RELATED WORK}
Forecasting future motion of nearby agents has been a topic of extensive study in recent years as the domain of autonomous transportation gained exponential interest both from numerous car manufacturers as well as various research institutions. Some of more traditional approaches to short-term motion prediction assume that the evolution of an object's state through time remains primarily governed by laws of physics and can be therefore modeled with e.g. well known \textit{bicycle model} \cite{gillespie1992fundamentals}. In addition, noise reduction techniques implementing filters, e.g. \textit{Kalman Filter} \cite{kalman1960new}, are often combined with the motion model to improve prediction accuracy and account for noise in measurements from sensors. Nevertheless, techniques such as the one presented in \cite{lin2000vehicle} are greatly restricted to simple environments and to very short prediction horizon (for instance 1 second into the future). Alternatively, models based on Bayesian Framework such as \textit{Bayesian Networks} (BNs) \cite{jensen1996introduction} and \textit{Hidden Markov Models} (HMMs) \cite{rabiner1989tutorial} are often employed to further account for the probabilistic nature of the problem and capture relationships between sets of random variables. For instance, \cite{schreier2014bayesian} models a probability distribution over a discrete set of maneuvers and then samples a number of possible future trajectories with respect to detected maneuvers. For a more comprehensive review of similar techniques we direct the reader to \cite{lefevre2014survey}. 

More recent and advanced techniques on the other hand focus on employing \textit{Deep Learning} (DL) \cite{goodfellow2016deep} based methods in order to account for both temporal and spatial complexity of the task. Lee \textit{et al.} introduced \textit{DESIRE} \cite{lee2017desire}, a framework for a multi-modal trajectory prediction based on CNNs, \textit{Conditional Variational Auto-Encoders} (CVAE) \cite{sohn2015learning}, and \textit{Recurrent Neural Networks} (RNN) \cite{rumelhart1986learning} encoder-decoder to predict multiple plausible future trajectories based on the encoded scene context, past motion histories, as well as interactions between multiple agents. Moreover, \cite{Rhinehart_2019_ICCV} used a generative model to predict agents' interactions from image and LiDAR data to then forecast plausible motion from joint state that encapsulates an encoding of all agents within the scene. Furthermore, the use of raster top-down view of scene context was presented in \cite{cui2019multimodal} with a focus on predicting multiple future paths as well as their associated probabilities through use of CNN. Another body of work that explores motion forecasting through probabilistic nature in multi-agent scenarios was demonstrated in \cite{tang2019multiple}. Authors modeled a joint behavior of numerous agents with a dynamic attention-based state encoder that allowed to capture past and future interactions for more precise multi-predictions. An interesting study was conducted by Skrikanth \textit{et al.} in \cite{srikanth2019infer} where instead of using a single top-down view image of the scene, multiple intermediate representations of objects were created by fusing images from either a stereo camera or LiDAR with depth information. Generated images represented lanes, roads, obstacles etc, and were used as an input to a CNN and LSTM \cite{hochreiter1997long} based network to obtain a future location of the vehicle of interest on an occupancy grid map. Recently proposed \textit{CoverNet} \cite{phan2020covernet} approaches the problem of motion forecasting by formulating it as classification problem over a diverse set of trajectories. Additionally, trajectories that were physically infeasible were excluded from the train set, thus, during inference if the model predicted such motion it would assign it a low probability, effectively restricting it from being taken into consideration. In \cite{marchetti2020mantra} Marchetti \textit{et al.} presented \textit{MANTRA}, a model based on memory networks \cite{weston2014memory} that learns the association between past and future motion and memorises most meaningful samples. Moreover, MANTRA presented capabilities of updating its internal representation of motion samples in an online fashion, thus allowing for continual improvement as new samples are collected. On the other hand, the following work \cite{gao2020vectornet} investigated a novel approach of transforming input features such as lanes, agents’ trajectories, and crosswalks into a vectorized form and incorporating those into a model based on \textit{Graph Neural Networks} (GNN) \cite{battaglia2018relational} called \textit{VectorNet}. The structure of VectorNet was divided into local and global graphs in order to allow the model to capture data from both individual polylines as well as from its aggregated global representation. Experimental results indicate a significant performance boost for predictions of up to 3 seconds into the future when compared with a CNN model based on ResNet-18 architecture, whilst reducing number of learnable parameters by 70\%.

\begin{figure*}[t]
  \centering
   \includegraphics[scale=.8]{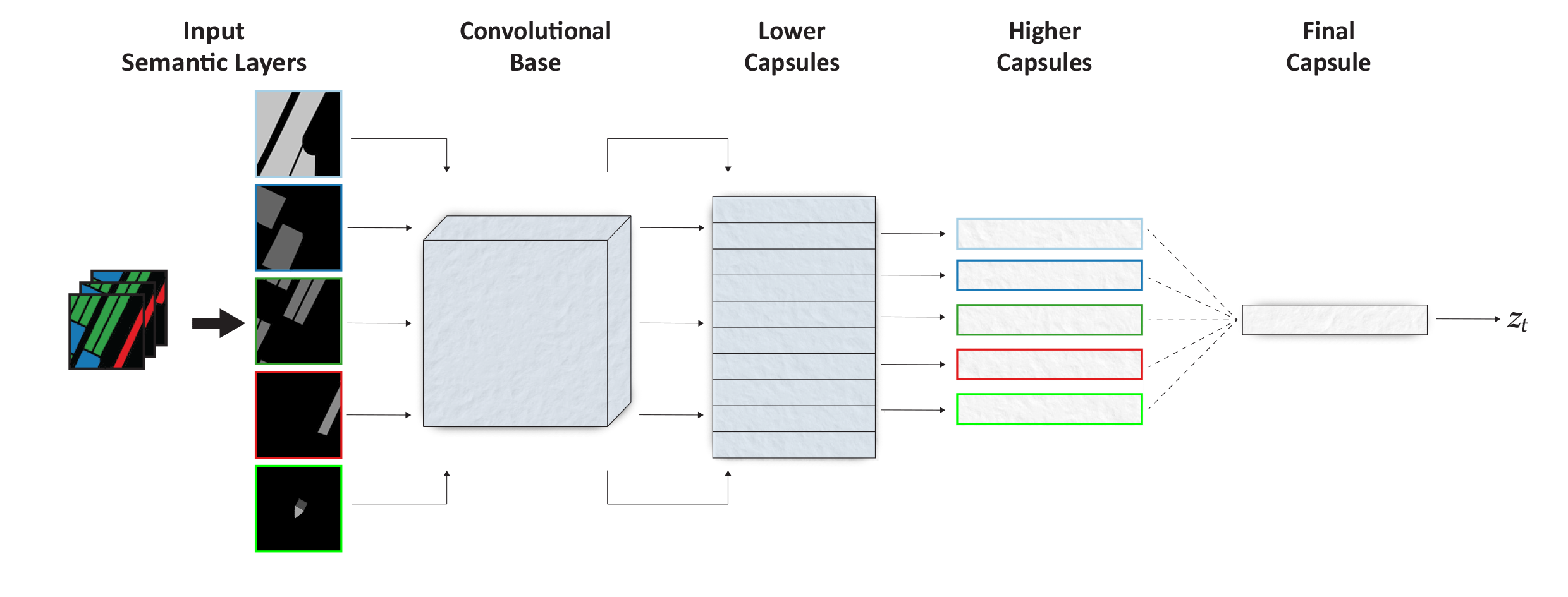}
  \caption{An overview of the proposed Capsule Net encoder. A single map chunk at time $t$ is disentangled into separate geometrical layers. Each layer is then passed separately to convolutional base, followed by lower capsules and then to its corresponding higher capsule. Finally, outputs of all higher capsules are concatenated and passed to final capsule to compute final encoding.}
  \label{fig:encoder_arch}
\end{figure*}
\section{PROPOSED APPROACH}
\subsection{Problem Formulation}
First, we assume access to a sufficient detection and tracking module that yields data corresponding to the current state of the tracked agent (e.g. position, velocity) at a fixed interval e.g. 2Hz. We also assume access to a HD map that define the following road layers; road segments, drivable areas, lanes, and walkways.

Furthermore, we are interested in predicting a sequence of an agent's future positions for $\tau$ time-steps into the future which we denote as a matrix $\mbox{\boldmath$\hat{Y}$} = [\mbox{\boldmath$\hat{y}$}_{t+1}, \mbox{\boldmath$\hat{y}$}_{t+2}, \cdots, \mbox{\boldmath$\hat{y}$}_{t+\tau}]$ where $\mbox{\boldmath$\hat{y}$}_{t+1}$ corresponds to a vector containing future predictions at time $t+1$ such that $\mbox{\boldmath$\hat{y}$}_{t+1} = (x_{t+1}, y_{t+1})$. Moreover, let $\mbox{\boldmath$ S$} = [\mbox{\boldmath$s$}_{t-\rho}, \cdots, \mbox{\boldmath$s$}_{t-1}, \mbox{\boldmath$s$}_{t}]$ denote a standardised matrix of the tracked agent's state features for past $\rho$ time-steps where:
\begin{equation}
    \mbox{\boldmath$s$}_t = [\mbox{\boldmath$v$}_t, \mbox{\boldmath$a$}_t, \theta_t]
    \label{eq:s_p}
\end{equation}
is a vector of features at time-step $t$ containing velocity $(x, y)$, acceleration $(x,y )$ and yaw angle respectively. In addition, we define a normalised tensor $\mbox{\boldmath$\mathsf M$} = [\mbox{\boldmath$\mathsf L$}_{t-\rho}, \cdots, \mbox{\boldmath$\mathsf L$}_{t-1}, \mbox{\boldmath$\mathsf L$}_{t}]$ which encapsulates local map chunks for past $\rho$ time-step. A single map chunk $\mbox{\boldmath$\mathsf L$}_t$ at time-step $t$ is further disentangled into separate layers (Fig. \ref{fig:map_dis}) such that $\mbox{\boldmath$\mathsf L$}_t = [\mbox{\boldmath$L$}_{t,0}, \mbox{\boldmath$\ L$}_{t,1}, \cdots, \mbox{\boldmath$ L$}_{t,n}]$ where $n$ is equal to number of layers that describe disentangled map, and $\mbox{\boldmath$ L$}_{t,i}$ is a matrix representing a single sparse image of a layer of type $i$. We use the same number of past time-steps $\rho$ for both \mbox{\boldmath$\mathsf S$} and \mbox{\boldmath$\mathsf M$} so that states as well as corresponding maps can be jointly encoded in a temporal manner.

\subsection{Sparse Semantic Layers}
In the literature, as previously outlined \cite{choi2019drogon, cui2019multimodal, marchetti2020mantra, phan2020covernet}, a common technique to obtain the context of the environment is to rasterise a large portion of its surrounding (e.g. $60m \times 60m$) into a top-down view HD map, allowing the network to detect and encode salient features as well as enable the agent to 'see' what lies in front of it. In contrast, we focus on creating a more meaningful encoding with respect to a single agent by exploiting its spatial representation within an environment through a semantic, hierarchical view of a local chunk of a given HD map. The following summarises the process of creating disentangled map chunks as depicted in Fig. \ref{fig:map_dis}.

First, let $\{(x_t, y_t) \mid \mbox{\boldmath$p$}_t \in \mbox{\boldmath$P$}\}$ denote a set of tracked agent's observed coordinates from time $t$ to $t-\rho$ where each $\mbox{\boldmath$p$}_t \in \mbox{\boldmath$P$}$ is used to define the origin of extraction for a map chunk $\mbox{\boldmath$\mathsf L$}_t$ at time $t$. Moreover, as previously mentioned, we define a single map chunk $\mbox{\boldmath$\mathsf L$}_t$ by $n$ sparse matrices (all pixels but the ones corresponding to the given layer are set to zero) $[\mbox{\boldmath$L$}_{t,0}, \mbox{\boldmath$\ L$}_{t,1}, \cdots, \mbox{\boldmath$ L$}_{t,n}]$ where the $\mbox{\boldmath$L$}_{t,i}$ corresponds to a semantic layer of type $i$. Thus, we can further extract each local semantic layer with respect to $\mbox{\boldmath$p$}_t$ as:
\begin{equation}
    \mbox{\boldmath$L$}_{t,i} = \Phi_{\mathrm{get\_l}}(\mbox{\boldmath$p$}_t, \lambda, \mathsf{s}), \quad \forall\ \mathsf{s} \in \mathsf{S} 
    \label{eq:layer_ext}
\end{equation}
where $\lambda$ defines the extraction offset in meters. We set the $\lambda=10$ and use a resolution of $3$ pixels per meter. The local chunk of the layer of interest is then extracted from $x_t - 3\lambda$ to $x_t + 3\lambda$ in a horizontal direction, and from $y_t - 3\lambda$ to $y_t + 3\lambda$ in a vertical direction, hence, the final size of the $\mbox{\boldmath$L$}_{t,i}$ is $20m \times 20m$. In addition, we create one extra layer that portrays the agent by rendering it at its origin defined by $\mbox{\boldmath$p$}_t$, with its initial orientation facing up. We then compute:
\begin{equation}
    \hat{\theta} = ((\pi / 2) + \mathrm{sign}(-\theta) \cdot \lvert \theta \rvert) \cdot 180 / \pi
    \label{eq:agent_theta}
\end{equation}
with respect to agent's $\theta$ which defines the rotation angle by which the agent is rotated so that its orientation is aligned with its global heading direction. We notice no improvement when layers, as well as the agent, are rotated so that their orientation aligns with agent’s heading facing up, and therefore decide to leave the original orientation with respect to the global coordinate system. Lastly, we transform extracted images (layers) from RGB to grayscale, and upscale each image so that its final size is equal to $64 \times 64$ pixels.

\subsection{Capsule Encoder's Architecture}
A typically employed approach that is used to encode the rasterised HD map is to use some form of pre-trained CNN architecture e.g. ResNet-50 that 1) does not provide equivariance with respect to extracted features and 2) uses pooling operations, effectively discarding useful information as a trade off for a small local translation invariance. With regards to the extraction of salient map features it is crucial to retain valuable data and encode hierarchical relationship between low and high level features. We propose to remedy these issues through the use of a spatial encoder that is based on the CapsNet as a feature extractor as opposed to the conventional CNN. We loosely follow the implementation from \cite{sabour2017dynamic}, and construct our CapsNet encoder as a four-part network as shown in Fig. \ref{fig:encoder_arch}.

Let $(k$, $s$, $o)$ denote the tuple that specifies a convolution layer's kernel size, stride size, and the number of output channels respectively (we set the padding value within all convolution layers to 0). First, we define a shallow convolutional base $\Phi_{\mathrm{base}}(\cdot)$ to extract local, low-level features of semantic layers which is composed of a single convolutional layer:
\begin{itemize}
    \item $(k=9, s=2, o=64)$
\end{itemize}
followed by the \textit{ELU} non-linearity \cite{clevert2015fast}. Next, we define the second part of the encoder i.e. lower level (primary) capsules $\Phi_{\mathrm{lower}}(\cdot)$ to learn parameters of more of a trivial parts of input data. Every capsule is a $4D$ unit where each of its dimension corresponds to the scalar output value of two consecutive convolutional operations:
\begin{itemize}
    \item $(k=9, s=2, o=32)$
    \item $(k=2, s=2, o=16)$
\end{itemize}
hence the number of convolution layers within $\Phi_{\mathrm{lower}}(\cdot)$ layer is equal to $4 \times 2$. We do not use any activation function in-between layers but rather run capsules' output vectors through the \textit{'squashing'} non-linearity (Eq: \ref{eq:squash}) \cite{sabour2017dynamic} in order to normalize the input vector \mbox{\boldmath$v$} so that the magnitude of short and long vectors is squashed to almost 0 and just below 1 respectively:
\begin{equation}
    \Phi_{\mathrm{squash}}(\mbox{\boldmath$v$}) = \frac{\|\mbox{\boldmath$v$}\|^2}{1+\|\mbox{\boldmath$v$}\|^2}\frac{\mbox{\boldmath$v$}}{\|\mbox{\boldmath$v$}\|}
    \label{eq:squash}
\end{equation}
Since low level features of input images e.g. edges and lanes resemble strong similarities among different types of semantic layers, it is reasonable to use a single $\Phi_{\mathrm{lower}}(\cdot)$ layer to learn and extract their parameters. Nonetheless, as their final representation differs significantly we decided to implement the subsequent part of the encoder by defining $n$ higher capsule layers $\Phi_{\mathrm{higher}}(\cdot)$ per each type of the input image. Hence, each type of semantic layer $\mbox{\boldmath$ L$}_{t,i}$ is encoded through its own respective higher capsule $\Phi_{\mathrm{higher}}(\cdot)$ that outputs a $32D$ vector of its latent representation. Lastly, we define the encoder's final part as a single capsule $\Phi_{\mathrm{final}}(\cdot)$ which outputs a $128D$ vector containing jointed representation of all semantic layers.

\subsection{Overall Framework}
We now describe the computational flow of data through the proposed network that yields short-term future positions for the agent of interest. Although the following represents computation process of a single agent, it is trivial to extend the method for a multi-agent scenario. For the purpose of simplicity and to focus on the main aspects of the paper we maintain a case of a single-agent.

\subsubsection{Encoding of Semantic Layers}
We transform each disentangled grayscale layer $\mbox{\boldmath$ L$}_{t,i}$ at time $t$ separately through the capsule encoder to compute its output representation vector by first running the matrix through convolutional base:
\begin{equation}
    \mbox{\boldmath$\mathsf Z_{t,i}$} = \Phi_{\mathrm{base}}(\mbox{\boldmath$ L$}_{t,i}) 
    \label{eq:conv_base}
\end{equation}
where \mbox{\boldmath$\mathsf Z_{t,i}$} is the output tensor containing $64$ feature maps of size $28 \times 28$. Next, the \mbox{\boldmath$\mathsf Z_{t,i}$} is further passed into the $ \Phi_{\mathrm{lower}}(\cdot)$ layer to compute lower capsules:
\begin{equation}
    \mbox{\boldmath$ Z_{t,i}$} =  \Phi_{\mathrm{lower}}(\mbox{\boldmath$\mathsf Z_{t,i}$}) 
    \label{eq:l_caps}
\end{equation}
where \mbox{\boldmath$ Z_{t,i}$} is the squashed output matrix with $400 \times 4D$ capsules (second convolution layer outputs $16$ feature maps of size $5 \times 5$). Furthermore, we run \mbox{\boldmath$ Z_{t,i}$} through its respective $ \Phi_{\mathrm{higher\_i}}(\cdot)$ to get the $32D$ output vector \mbox{\boldmath$ z_{t,i}$} whose scalar values corresponds to the layer's $\mbox{\boldmath$ L$}_{t,i}$ latent instantiation parameters:
\begin{equation}
    \mbox{\boldmath$ z_{t,i}$} =  \Phi_{\mathrm{higher\_i}}(\mbox{\boldmath$Z_{t,i}$}) 
    \label{eq:h_caps}
\end{equation}
We repeat this process for each semantic layer and concatenate each output to create an input matrix for the final capsule:
\begin{equation}
    \mbox{\boldmath$ z_{t}$} =  \Phi_{\mathrm{final}}(\mathrm{concat}(\mbox{\boldmath$z_{t,0}$}, \mbox{\boldmath$z_{t,1}$}, \cdots, \mbox{\boldmath$z_{t,n}$}))
    \label{eq:h_caps}
\end{equation}
that outputs a final $128D$ vector \mbox{\boldmath$ z_{t}$} at time $t$.

\subsubsection{Encoding-Decoding of Future Motion} Moreover, each state vector \mbox{\boldmath$ s_{t}$} is encoded through a fully-connected layer with $128$ output units, followed by the ELU activation. We then concatenate it with a corresponding \mbox{\boldmath$ z_{t}$} to form the matrix of shape $[\rho, 256]$ where $\rho$ refers to the number of observed time-steps. The matrix is then encoded in temporal manner through use of the LSTM layer. We set the size of the layer's hidden-state to $128$. Finally, we run the last output of the LSTM through the decoder (fully-connected) layer with $2 (x, y) \times \tau$ units where $\tau$ corresponds to the future time-horizon, to get future predictions. The output of the decoder can be reshaped to create the target matrix $\mbox{\boldmath$\hat{Y}$} = [\mbox{\boldmath$\hat{y}$}_{t+1}, \mbox{\boldmath$\hat{y}$}_{t+2}, \cdots, \mbox{\boldmath$\hat{y}$}_{t+\tau}]$ with each vector containing predicted future position $(x, y)$ of the tracked agent, relative to its position at the last observed time-step $t$.

\begin{table*}[bp]
\caption{Ablation study with different backbone architectures. Each cell contains ADE/FDE error in meters for given future time horizon. In addition we examine whether additional map data, collected over observed time-steps $\rho$ provides performance increase. Number of each backbone parameters is reported in millions.}
\label{tb:cnn_caps_encoder}
    \begin{center}
    \setlength{\extrarowheight}{3pt}
      \begin{tabular}{
        ||l||
        >{\centering\arraybackslash}p{1.25cm}| >{\centering\arraybackslash}p{1.25cm}|
        >{\centering\arraybackslash}p{1.25cm}| >{\centering\arraybackslash}p{1.25cm}|
        >{\centering\arraybackslash}p{1.25cm}| >{\centering\arraybackslash}p{1.25cm}|
        >{\centering\arraybackslash}p{1.5cm}||
        }
        \hline
        \multirow{2}{*}{\shortstack{Backbone}} & 
        \multicolumn{6}{c|}{Future Time Horizon (seconds)} &%
        \multirow{2}{*}{\shortstack{\shortstack{\#Params\\(Backbone)}}} \\ 
        \cline{2-7}
         & 1s & 2s & 3s & 4s & 5s & 6s & \\
        \hline
        \hline
        ResNet-50$_1$            & 0.25/0.32 & 0.72/1.29 & 1.22/2.45 & 1.41/3.39 & 2.18/5.46 & 2.93/7.59 & 23.5m \\
    
        ResNet-50$_{\rho}$       & 0.22/0.30 & 0.48/0.90 & 1.11/2.16 & 1.39/3.28 & 2.10/5.15 & 2.82/7.12& 23.5m\\
        \hline
        Capsule-Encoder$_1$      & \textbf{0.20/0.29} & 0.47/0.89 & 0.89/1.90 & 1.44/3.32 & 2.13/5.17 & 2.95/7.25 & 0.95m\\
        
        Capsule-Encoder$_{\rho}$ & \textbf{0.20/0.29} & \textbf{0.46/0.88} & \textbf{0.84/1.85} & \textbf{1.34/3.17} & \textbf{1.99/4.91} & \textbf{2.74/6.89} & 0.95m\\

        \hline
        \end{tabular}
    \end{center}
\end{table*}
\section{EXPERIMENTS AND RESULTS}

\subsection{Dataset}
We report results of our experiments on the publicly available self-driving dataset \textit{nuTonomy Scenes} (nuScenes) \cite{caesar2020nuscenes} designed for variety of tasks such as detection, tracking as well as motion prediction. nuScenes provides access to 1000 scenes (approximately 20 seconds each) that were collected in Boston and Singapore. Collected scenes provide a wide diversity with regards to weather conditions, traffic situations and traffic density. In addition, nuScenes contains human-annotated vectorized maps with 11 different semantic layers. Scenes and objects within each scene (e.g. vehicles, pedestrians) were accurately annotated at the rate of $2Hz$ and modeled as a cuboid, providing further access to object's position, size and yaw angle. We notice that some parts of the HD map with respect to agents' positions are not rendered properly and we therefore remove these samples. 

\subsection{Training Setup and Evaluation Metrics}
The proposed network is trained for $70$ epochs with \textit{Adam} optimizer \cite{kingma2014adam} and set the initial learning rate to $5e-4$. We then reduce the learning rate at epoch $5$ and $20$ by $\gamma=0.1$. We optimise the network by minimising:
\begin{equation}
    \mathcal{J} = \alpha \mathcal{J}_1 + \beta \mathcal{J}_2
    \label{eq:loss}
\end{equation}
where the $\mathcal{J}_1$ is the mean absolute error and the $\mathcal{J}_2$ is the mean squared error. We notice a slight improvement when combining both losses, and through empirical evaluation we find that setting $\alpha = 1.0$ and $\beta = 1.0$ provides best results. Dataset is further split in accordance to nuScene's $train$ and $val$ split sets \footnote[1]{Available on the official repository}. Since the $test$ set has not been annotated we split the $train$ set into train and validation sets and use the $val$ split as a test set. For every sample, we observe $2$ seconds of agent's past states and make predictions for every $t \in \{1, 2, 3, 4, 5, 6\}$ seconds of the future time horizon. We report quantitative results by employing \textit{Average Displacement Error} (ADE) between all predictions, and \textit{Final Displacement Error} (FDE) between final prediction at time $t$. Prediction errors for both ADE and FDE is measured in meters. In addition, we include the comparison with respect to the number of parameters within adopted feature extractors (e.g. ResNet-50) as well as networks as a whole. 

\subsection{Ablation Study - CNN vs Capsule based feature extractor}
First, we analyse the performance of the network by employing two backbone feature extractors i.e. 1) our proposed Capsule based encoder, and 2) a commonly used ResNet-50 architecture. In addition, as previously discussed, the encoder uses disentangled map chunks that are collected over observed past $\rho$ times-steps, we therefore conduct a further study on the impact of incorporating these map chunks over time against the use of a single map chunk obtained at the initial time $t$. We notice that the initial, large output size of ResNet-50 (i.e. $2048$ units) leads to either significant overfitting or exploding gradient early in the training. To account for this, we downsample the output through additional single fully-connected layer to $128$. Results are presented in Table \ref{tb:cnn_caps_encoder} with each cell containing the ADE/FDE error for the corresponding time horizon. As demonstrated, the proposed Capsule encoder whose input considers multiple map chunks over time (Capsule-Encoder$_\rho$) outperforms other methods. Interestingly, for the first 3 seconds, there is no significant improvement between Capsule-Encoder$_\rho$ and Capsule-Encoder$_1$ (single map at time $t$), however, as the model is looking further into the future, the performance difference becomes more apparent. Next, both Capsule-Encoder and ResNet-50 demonstrate an improved performance when spatial data for all observed time-step $\rho$ is included. Lastly, it is crucial to emphasize that apart from significant performance improvement over ResNet-50 backbone, the Capsule encoder is substantially lighter, having less than 1 million parameters as opposed to ResNet-50 (over 23 million).
\begin{table*}[bp]
\caption{Comparison study between our final model vs two physics-based models and two recently proposed methods from the literature. Again, we provide a ADE/FDE error for each second of the future time horizon for up to 6 seconds. For CoverNet$^{\epsilon}$ (modes) we adjust number of modes with accordance to the $\epsilon$ value. We also included number of parameters (millions) of each model.}
\label{tb:sota_comparison}
    \begin{center}
    \setlength{\extrarowheight}{3pt}
        \begin{tabular}{
        ||l||
        >{\centering\arraybackslash}p{1.25cm}| >{\centering\arraybackslash}p{1.25cm}|
        >{\centering\arraybackslash}p{1.25cm}| >{\centering\arraybackslash}p{1.25cm}|
        >{\centering\arraybackslash}p{1.25cm}| >{\centering\arraybackslash}p{1.25cm}|
        >{\centering\arraybackslash}p{1.5cm}||
        }
        \hline
        \multirow{2}{*}{\shortstack{Model}} & 
        \multicolumn{6}{c|}{Future Time Horizon (seconds)} &%
        \multirow{2}{*}{\shortstack{\#Params}} \\ 
        \cline{2-7}
         & 1s & 2s & 3s & 4s & 5s & 6s & \\
        \hline
        \hline
        
        Const. Vel. \& Head.    & 0.48/0.66 & 0.96/1.75 & 1.60/3.32 & 2.38/5.30 & 3.28/7.61 & 4.28/10.22 & N/A\\
    
        Physics Oracle      & 0.42/0.55 & 0.77/1.35 & 1.26/2.55 & 1.89/4.18 & 2.64/6.15 & 3.50/8.44 & N/A\\
        \hline
        CoverNet$^8$ (64)  \cite{phan2020covernet} & 0.74/0.96 & 1.22/1.98 & 1.82/3.38 & 2.33/4.93 & 3.24/7.10 & 4.07/8.89 & 32.0m\\
        
        CoverNet$^4$ (415) \cite{phan2020covernet} & 0.73/0.94 & 1.18/1.89 & 1.76/3.31 & 2.25/4.68 & 3.15/6.89 & 4.09/9.34 & 33.5m\\
    
        CoverNet$^2$ (2206) \cite{phan2020covernet} & 0.65/0.86 & 1.13/1.90 & 1.73/3.32 & 2.23/4.63 & 3.24/7.23 & 4.23/9.74 & 40.9m\\
        
        MTP \cite{cui2019multimodal} & 0.46/0.63 & 0.81/1.36 & 1.29/2.57 & 1.79/3.97 & 2.49/5.83 & 3.44/8.28 & 32.0m\\
        \hline
        Our Final Model  & \textbf{0.20/0.29} & \textbf{0.46/0.88} & \textbf{0.84/1.85} & \textbf{1.34/3.17} & \textbf{1.99/4.91} & \textbf{2.74/6.89} & 1.2m\\
        \hline
        \end{tabular}
    \end{center}
\end{table*}

\subsection{Comparison with state-of-the-art methods}
Finally, we compare our results with two physics-based models as well as with two state-of-the-art approaches that use CNN based on ResNet-50 as well as the rasterised top-down view of the surrounding:
\begin{itemize}
    \item \textbf{Constant Velocity \& Heading Angle}: Dynamic model explained in \cite{phan2020covernet}, based on an assumption of the agent having constant velocity and heading angle across the prediction time.
    \item \textbf{Physics Oracle}: An extended version of physics-based models explained in \cite{phan2020covernet}. This model computes future prediction for numerous physics-based models and chooses the one with the minimum $L2$ distance with respect to the ground truth data.
    \item \textbf{MTP}: This is the implementation of the model presented in \cite{cui2019multimodal}. In the original paper, authors focused on multi-modal prediction for fixed amount of modes. Since our model focuses on deterministic prediction, we train the MTP to predict a single mode. We adjust the learning settings with accordance to details provided within the paper.
    \item \textbf{CoverNet:} Another method proposed in \cite{phan2020covernet} for multi-modal prediction. We examine this model across three different settings of error tolerance $\epsilon$ (2, 4, 8) that were publicly provided by authors (for detailed information refer to the original publication). CoverNet was trained to predict probability distribution over a set of diverse trajectories, we therefore use the trajectories that were provided for each of the error tolerance and pick top$^1$ prediction as the final trajectory. Again, as with MTP we also adopt training settings for CoverNet with accordance to details outlined in the original paper.
\end{itemize}
Results of the comparison study are presented in Table \ref{tb:sota_comparison}. It can be seen that our approach based on CapsNet significantly outperforms all other methods whilst drastically reducing number of parameters within the network. A noticeable difference between our approach vs MTP and CoverNet might further imply that using, for instance ResNet based backbone could not be considerably beneficial due to e.g. large output size which, when re-training these networks, we noticed leads to early overfitting. Furthermore, pre-trained ResNet-50 consists of weights that were optimised for a completely different domain \cite{deng2009imagenet}, and therefore it is arguable whether such an approach leads to a valuable and meaningful extraction of salient features from the global HD map.

\section{DISCUSSION}
In this work we explored a novel approach to learning salient features from local HD maps. In essence, our work investigated how a spatial encoder based on the CapsNet architecture can be used to replace a generally employed CNN that uses a pre-trained backbone model e.g ResNet-50 as a feature extractor. In addition, we propose a use of local HD map chunks as opposed to global maps, that can be further disentangled into separate, semantic geometrical layers to encode more meaningful latent representation of a tracked agent. Experiments on public dataset (nuScene) presents satisfactory results with respect to prediction of a deterministic trajectory. Moreover, our model outperforms recently proposed methods from the literature, whilst substantially reducing size of the network. In the future, we would like to extend this approach to multi-modal predictions in order to account for inherently uncertain future that is associated with the discussed task.

\addtolength{\textheight}{0cm}   

\bibliographystyle{plain}
\bibliography{root}
\end{document}